\documentclass[journal]{IEEEtran}
\usepackage{cite}
\usepackage{amsmath,amssymb,amsfonts}
\usepackage{algorithmic}
\usepackage{graphicx}
\usepackage{algorithm}
\usepackage{textcomp} 
\usepackage{hyperref}
\def\BibTeX{{\rm B\kern-.05em{\sc i\kern-.025em b}\kern-.08em
    T\kern-.1667em\lower.7ex\hbox{E}\kern-.125emX}}
\begin{document}
\title{ ME\textsuperscript{2}FDiff: Multiscale Expert Enhancing Feature Diffusion in Residual Network for Brain Tumor Segmentation}

\author{Afrida Rahman and Bijoy Kundu
\thanks{This work was supported in part by the Ivy Biomedical Innovation Funds, start-up funds from the University of Virginia, and Commonwealth Commercialization Funds from the Virginia Innovation Partnership Corporation. The associated technology is protected under U.S. Patent Application No. 19/720,872 (June 25, 2026).}
\thanks{ Afrida Rahman is with the Department of Radiology and Medical Imaging, University of Virginia, VA, USA (e-mail: nyp9tp@virginia.edu). }
\thanks{Bijoy Kundu is with the Department of Radiology and Medical Imaging, Biomedical Engineering, School of Data Science, University of Virginia, VA, USA (e-mail: bkk5a@virginia.edu).}}

\maketitle

\begin{abstract}
Automated segmentation of post-treatment glioma from MRI is essential for accurate diagnosis and treatment planning, yet remains difficult. Surgery and chemoradiotherapy cause structural changes in brain like resection cavities, edema, and fibrosis, that alter imaging appearance and mimic tumor recurrence. Contemporary segmentation architectures attain high accuracy with two limitations. First, the bottleneck originates from the deepest encoder stage, so one representation must serve sub-regions that differ widely in size and appearance. Second, they produce the segmentation in a single forward pass with no mechanism for refinement. We propose ME$^2$FDiff, a single-model architecture, addressing both. An attention-conditioned spectral diffusion block refines the bottleneck. It uses the image-conditioned latent diffusion process with two processes: a spatial pre-attention gate and a spectral fusion block whose timestep-adaptive bandpass filter selects the informative frequency band at each noise level. A multiscale expert-enhanced feature block fuses the intermediate encoder scales into the bottleneck through expert-selected shared tokens, guaranteeing balanced utilization without an auxiliary loss. On the BraTS 2024 post-treatment glioma dataset, ME$^2$FDiff achieves the best results on every tumor sub-region among all single-model baselines (91.23\% Dice, 84.52\% IOU, 2.13 mm HD95 on whole-tumor) and surpasses a three-expert ensemble on both overlap metrics at roughly one-third of its inference cost. Against recent state-of-the-art methods retrained under identical conditions, it maintains the strongest performance with the fewest parameters of any competing model. These findings demonstrate that a single, compact network can absorb the specialization, usually sought through ensembling or larger architectures.
\end{abstract}


\section{Introduction}
\label{sec:introduction}
\IEEEPARstart{G}{liomas} are a prevalent malignant brain tumor and a major cause of cancer-related mortality in adults~\cite{price2024cbtrus}. Their infiltrative and diffusive nature yields poorly defined borders, which makes segmentation difficult. Manual delineation is labor-intensive and subject to inter-observer variability~\cite{menze2014multimodal}. Magnetic resonance imaging (MRI) is the standard modality for glioma evaluation~\cite{baig2024segmentation}. While pre-treatment scans show clearly enhanced tumor and edema~\cite{ranjbar2023towards}, post-treatment scans are more complex, exhibiting radiation necrosis, resection cavities, and post-surgical inflammation that often resemble recurrence. Benchmark challenges like BraTS have accelerated advances in automated segmentation, with BraTS2024 focusing on the post-treatment setting~\cite{de20242024}. Recent encoder-decoder networks~\cite{hatamizadeh2022swinunetr, yao2024cnn, isensee2020nnunetbrats} achieve high aggregate performance but share two limitations. First, the bottleneck originates only from the deepest encoder stage. The shallower stages hold the fine detail needed to find small structures, but they reach the decoder only through skip connections and never enter the bottleneck itself. As a result, one feature representation has to describe every subregion at once, even though they differ widely in the large, small, and sharply outlined regions. Second, the network predicts the segmentation in a single pass and has no way to correct it
afterwards. This is a problem in post-treatment scans, where the boundary between tumor and treatment effects is often unclear, and a first prediction is easy to get wrong. Ensembling is the established remedy for the first limitation ~\cite{10.1007/3-540-45014-9_1}. Several architectures are trained independently, and their predictions are combined so that the strengths of one compensate for the weaknesses of another. However, it is quite expensive; inference increases with the number of experts, and the combination is usually controlled by a predetermined mechanism, such as a sequential cascade, uniform averaging, or confidence-derived weighting, none of which can reason about the input. Diffusion models naturally address the second limitation by replacing a single deterministic forward pass with an iterative denoising process that progressively refines predictions ~\cite{yuan2026discrete}. However, they are computationally prohibitive at volumetric resolution when used naively in image space.

We introduce a single volumetric segmentation network, ME$^2$FDiff, that meets these two limitations without recourse to multiple networks. The design follows the routing mechanism that arbitrates between separate models in an ensemble, which is equally applicable to arbitrating between encoder scales inside a single network. The diffusion process that is intractable over an image volume becomes affordable when applied to a compact mask latent space. A multimodal MRI volume is first encoded into multi-scale stage features and a deep bottleneck embedding. The Attention-Conditioned Spectral Diffusion (ACSDiff) module refines this bottleneck through an image-conditioned latent diffusion process, which is built by a spatial pre-attention gate (SPG) and a spectral diffusion conditioner (SDC). Here, the SPG sharpens the conditioning features, and the SDC module couples the diffusion-noise and semantic streams in the frequency domain under a timestep-adaptive band-pass filter. Before the result is decoded to the final segmentation, the multiscale expert-enhanced feature block (ME$^2$FB) fuses all intermediate encoder stages into the refined bottleneck through expert-selected routing. Therefore, our proposed network shows the strongest held-out test performance in its model family, with an average Dice of 0.857, at a fraction of its computational cost. The main contributions of this work are summarized as follows:
\begin{itemize}
\item We integrate a multiscale expert-enhanced feature block (ME$^2$FB) into a SegResNet-VAE backbone, the first use of expert routing within a VAE-regularized residual CNN for 3D brain tumor segmentation.
\item We propose ACSDiff, an attention-conditioned spectral diffusion that refines the encoder bottleneck in the frequency domain, coupled with a spatial pre-attention gate that sharpens its conditioning features.
\item The combination of these components within a single network, optimized by the hybrid loss functions (Dice, Tversky, and Focal), achieves the SOTA performance at the lowest parameter count.
\item Our method outperforms not only every individual backbone but also the multi-model ensembles assembled from state-of-the-art segmentation networks on BraTS2024 post-treatment gliomas.
\end{itemize}

\section{Related Work}
\label{sec:related}

Deep learning has driven successive breakthroughs in brain tumor segmentation~\cite{moeskops2016automatic, liang2022transconver}. Its strength is the ability to extract discriminative features directly from data without the need for manually created priors~\cite{kharaji2024brain}. Convolutional architectures established the dominant paradigm on the pre-treatment BraTS datasets~\cite{thillaikkarasi2019enhancement, wang2021ensemble}. Afterwards, the U-Net encoder-decoder became the standard backbone, enriched with attention, residual, and multi-scale mechanisms~\cite{maji2022attention, ronneberger2015unet}. Convolutions capture long-range dependencies only indirectly, which led to attention-based designs~\cite{liang2022transconver}. Several of which are used for our method comparison: TransUNet~\cite{chen2021transunet} applies a Vision Transformer to CNN-extracted features and decodes them with a U-Net; SegFormer~\cite{xie2021segformer} pairs a hierarchical encoder with no positional encoding and a lightweight MLP decoder; MedFormer~\cite{gao2022medformer} reduces the quadratic cost of self-attention on 3D data with bidirectional multi-head attention; U-Mamba~\cite{ma2024umamba} replaces attention with a state-space model that captures long-range context in linear time; and MoE-SAM~\cite{LiRuo_MoESAM_MICCAI2025} adapts the Segment Anything foundation model~\cite{kirillov2023sam} through a mixture-of-experts feature-enhancement stage. Despite their differences, these designs have one structural characteristic in common: the bottleneck is created by the deepest encoder stage alone, while shallower stages that carry the detail required to localize small sub-regions enter merely as skips.

To compensate for architecture-dependent variability, ensembling combines independently trained networks through majority voting, probability, or weighted averaging or stacking~\cite{10.1007/3-540-45014-9_1}. It dominates recent BraTS submissions, at the price of inference that scales with the number of experts. Mixture-of-experts routing~\cite{moeskops2016automatic} offers a learned alternative to such fixed combination rules, though it has served chiefly to combine models rather than to allocate capacity across scales within one. A separate line of work replaces the single deterministic forward pass with iterative refinement: denoising diffusion~\cite{ho2020ddpm} has been applied to segmentation by diffusing the mask conditioned on the image. However, this has been limited to 2D image space, where computing spatial attention affinity remains affordable but does not scale to 3D volumes. Our method draws on both by relocating each inside a single network that operates directly in 3D volume.

\section{Materials and Methods}
\begin{figure*}[h!]
    \centering
    \includegraphics[width=\linewidth]{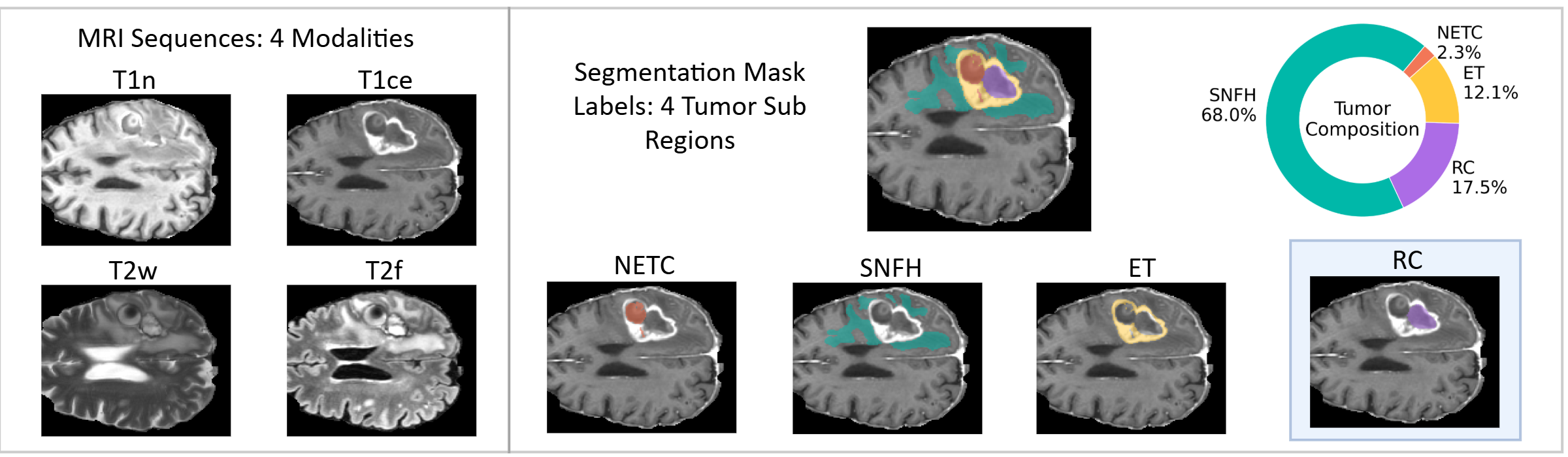}
    \caption{BraTS2024 dataset. The data set has four brain MRI scans i.e. pre-contrast T1-weighted (T1n), contrast-enhanced T1-weighted (T1ce), T2-weighted fluid-attenuated inversion recovery (T2f), T2-weighted (T2w) and four sub-regions in segmentation mask i.e. enhancing tissue (ET), surrounding non-enhancing T2/fluid-attenuated inversion recovery (FLAIR) hyperintensity (SNFH), non-enhancing tumor core (NETC), and the newly added tissue sub-region, resection cavity (RC).}
    \label{figure:1}
\end{figure*}

\subsection{Dataset}
\label{sec:dataset}
In this study, experiments are conducted on the BraTS 2024 Adult Glioma Post-Treatment (GLI) dataset, which, unlike earlier pre-operative BraTS editions, includes a resection cavity (RC) label and is thus representative of clinically realistic post-surgical glioma MRI. The dataset comprises $1{,}621$ preprocessed volumes drawn from $731$ unique patients, with multiple longitudinal timepoints contributing several volumes per patient. Each case provides four co-registered MRI modalities, i.e. native T1-weighted (T1n), contrast-enhanced T1-weighted (T1c), T2-weighted (T2w), and T2-FLAIR (T2f) and stacked in the fixed channel order $[\text{T1n}, \text{T1c}, \text{T2w}, \text{FLAIR}]$. Voxel-wise annotations are provided for five classes: background (0), necrotic core (NCR, 1), edema / surrounding non-enhancing FLAIR hyperintensity (ED/SNFH, 2), enhancing tumor (ET, 3), and resection cavity (RC, 4). Resection cavity is the newly added tissue subregion, which is necessary for those patients who are receiving adjuvant radiotherapy after surgical resection. \cite{de20242024} (\textbf{Figure \ref{figure:1}})\\

\subsubsection{Preprocessing}
All volumes are preprocessed using a fixed pipeline prior to model training. Each modality is intensity-normalized independently using $z$-score normalization computed only over non-zero (brain) voxels, so that statistics are not skewed by the zero-valued background; background voxels remain at zero after normalization. Initially, the BraTS MRI scans have large dimensions i.e. $182 \times 218 \times 182$. Volumes are subsequently cropped to the bounding box of the non-zero brain region to remove uninformative background. Later, it is resampled to a fixed isotropic grid of $128 \times 128 \times 128$ voxels using trilinear interpolation (order $1$) for the image channels and nearest-neighbor interpolation (order $0$) for the segmentation masks to preserve discrete label values. To standardize the 3D resolution, all the images are cropped using the ratios [54:184, 56:184, 13:141] throughout the three spatial dimensions. Since the validation data lacked ground truth labels, the entire training dataset of 1621 samples is divided into training and validation subsets using a 70:20:10 split. This results in 1134 samples for training, 324 samples for validation, and 163 samples for testing.

\subsection{Baseline Methods}
To situate the proposed model within the current state of the art, we establish two families of baselines of increasing strength. The first comprises individually trained segmentation architectures spanning the principal design paradigms in volumetric medical imaging: convolutional, transformer-based, and foundation-model  (Section~\ref{subsubsec:single_baselines}). The second comprises ensemble fusion strategies that aggregate the predictions of several such architectures at inference (Section~\ref{subsubsec:ensemble_baselines}).
\begin{figure*}[h!]
    \centering
    \includegraphics[width=\linewidth]{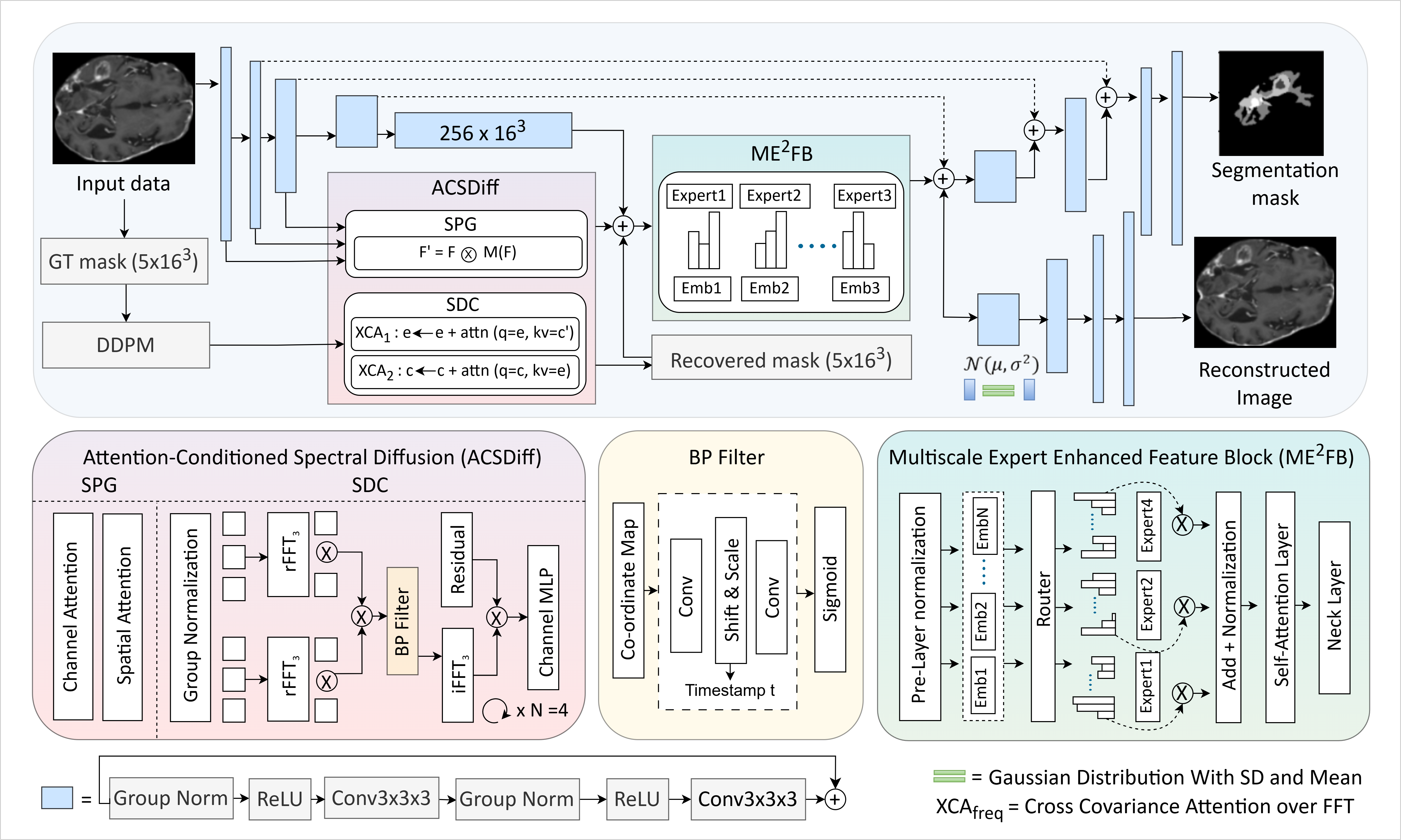}
    \caption{Overall model architecture of our proposed method. It has two important modules at the bottleneck of SegResNetVAE, i.e 1. Attention-Conditioned Spectral Diffusion (ACSDiff) combined with two modules: (a) Spatial Pre-attention Gate (SPG), (b) Spectral Diffusion Conditioner (SDC) with time-adaptive Band Pass Filter (NBP filter), 2. Multiscale Expert Enhanced Block (ME$^2$FB)}
    \label{figure:2}
\end{figure*}

\subsubsection{Single-model architectures} \label{subsubsec:single_baselines} 
A set of model architectures is implemented and trained independently on the four-channel (T1n, T1c, T2w, FLAIR) input. Among convolutional designs, the 3D U-Net~\cite{ronneberger2015unet} is the symmetric encoder-decoder that founded the paradigm, propagating high-resolution detail to the decoder through skip connections. SegResNetVAE~\cite{myronenko20183d,kundu2025ai} extends a residual encoder-decoder with a variational branch that reconstructs the input from the bottleneck, regularizing the encoder representation under limited training data. AHNet~\cite{liu20173danisotropichybridnetwork} transfers features learned from 2D pretraining into a 3D network with anisotropic kernels, decoupling in-plane from through-plane feature extraction. MedNeXt~\cite{roy2023mednext} adapts the ConvNeXt block to volumetric segmentation, combining depthwise convolutions with inverted bottlenecks and large kernels. Among transformer-based designs, UNETR~\cite{hatamizadeh2022unetr} replaces the convolutional encoder with a plain vision transformer applied to 3D patches. Swin UNETR~\cite{hatamizadeh2022swinunetr} instead employs a hierarchical encoder that applies self-attention within local windows shifted between successive layers, capturing long-range context at tractable volumetric cost. Finally, SAM introduces a promptable foundation model that generates segmentation masks from points, text prompts and generalizes to unseen objects without task-specific retraining~\cite{kirillov2023sam}.\\
 
\subsubsection{Ensemble Fusion Strategies}
\label{subsubsec:ensemble_baselines}
 
As no single architecture is uniformly optimal across sub-regions, ensembling constitutes the stronger baseline family. All strategies fuse the three best models, i.e., SegResNetVAE, SwinUNETR, and MedNeXt, selected based on their individual average Dice scores and HD95. Let $p_{k,c}(v)$ denote the probability assigned by expert $k$ to class $c$ at voxel $v$, and $\hat{p}_{c}(v)$ the fused prediction, decoded as $\hat{y}(v) = \arg\max_{c} \hat{p}_{c}(v)$. Dynamic weighting~\cite{guan2024dynamic} relaxes uniform averaging by letting each expert's contribution vary per voxel, with weights derived analytically from its own predictive confidence and normalized across experts; cascade fusion~\cite{wang2021ensemble} combines the experts sequentially rather than in parallel, stitching three independently trained stage networks at inference in the order WNet $\rightarrow$ TNet $\rightarrow$ ENet, so that each stage refines only within the region delineated by its predecessor. Finally, \emph{MoE gating}~\cite{moeskops2016automatic} instead learns the combination, using a gating network trained over the frozen experts to predict input-conditional weights. These three strategies (cascaded refinement, learning gate, and confidence weighting) are evaluated as ensemble baseline methods against our proposed work.

\subsection{Proposed Method}
\label{sec:proposed}
\subsubsection{Overview}
\label{subsec:overview}

The architecture of ME$^2$FDiff is illustrated in Fig.~\ref{figure:2}. A multi-modal MRI volume is encoded by a SegResNetVAE encoder into a hierarchy of multi-scale stage features together with a deepest bottleneck embedding $c^{0} \in \mathbb{R}^{256 \times 16^{3}}$, which serves as the semantic condition. This representation is refined by two modules in sequence. The first, {ACSDiff}, comprises a Spatial Pre-attention Gate ({SPG}) that sharpens the conditioning features in the image domain. It is coupled with a Spectral Diffusion Conditioner ({SDC}) that fuses the diffusion noise and semantic streams in the frequency domain under a timestep-adaptive band-pass gate, and the pre-attended features condition the diffusion directly. The second module, Multiscale Expert Enhanced Feature Block ({ME$^2$FB}), then rebuilds the refined bottleneck from all encoder scales, using expert-choice routing to determine which scale-specific evidence is emphasized at which spatial location. The resulting representation is returned to the SegResNetVAE decoder, which produces the final segmentation. The two modules are trained jointly under a composite objective comprising a latent noise-prediction term $L_{\epsilon} = \lVert \epsilon - \hat{\epsilon} \rVert_{2}^{2}$\quad following DDPM~\cite{NEURIPS2020_4c5bcfec}, an auxiliary segmentation term $L_{\text{seg}}$, a class-weighted combination of Tversky, Dice, and Focal losses applied to the decoded
prediction and the variational regularizer $L_{\text{vae}}$ of the backbone:

\begin{equation}
L_{\text{total}} = L_{\epsilon} + \lambda_{\text{seg}} L_{\text{seg}} + \lambda_{\text{vae}} L_{\text{vae}}.
\label{eq:total_loss}
\end{equation}

\subsubsection{Attention-Conditioned Spectral Diffusion}
\label{subsec:m1}

Module ACSDiff introduces such a mechanism by formulating mask refinement as an image-conditioned latent diffusion process and conditioning that process through two attention blocks acting in complementary domains: SPG, a spatial pre-attention gate applied to the features that supply the condition; and SDC, a spectral fusion block that couples the condition to the diffusion noise in the frequency domain.
\paragraph{Latent mask diffusion}
Diffusing the mask in image space is computationally prohibitive in three dimensions. ACSDiff instead operates on a compact latent: the ground-truth mask is compressed to a latent $z_{0} \in \mathbb{R}^{5 \times 16^{3}}$ and corrupted by the DDPM forward process to $z_{t}$, from which a denoiser predicts the added noise $\hat{\epsilon}$ conditioned on $c^{0}$. Operating on a low-resolution latent rather than in image space keeps the generative refinement tractable in three dimensions while retaining full volumetric condition. At inference, the mask is unavailable, and the latent is instead sampled from Gaussian noise and denoised by a conditional DDIM reverse process.

\paragraph{Spatial Pre-attention Gate (SPG)}
The condition is only as informative as the features from which it is formed. SPG therefore refines each encoder stage at its native resolution using a lightweight convolutional block attention module~\cite{woo2018cbam}, emphasizing informative channels and tumor-relevant voxels before the features are fused. Channel attention aggregates globally averaged and max-pooled descriptors through a shared reduction MLP followed by a sigmoid, while spatial attention convolves the channel-wise average and maximum maps into a single sigmoid gate over voxels:

\begin{equation}
F' = M_{s}\big(M_{c}(F) \otimes F\big) \otimes \big(M_{c}(F) \otimes F\big).
\label{eq:preattn}
\end{equation}
where $F$ and $F'$ are the input and refined stage features, $M_{c}(\cdot)$ and $M_{s}(\cdot)$ denote the channel and spatial attention maps, and $\otimes$ is element-wise multiplication with broadcasting. Being of linear complexity, SPG is affordable at the $128^{3}$, $64^{3}$, and $32^{3}$ stage resolutions, at which self-attention would be restrictive. Both the diffusion condition and the ME$^2$FB fusion consequently operate on saliency-reweighted features.

\paragraph{Spectral diffusion Conditioner (SDC)}
The denoiser must reconcile two embeddings of differing provenance: the noise embedding $e$, derived from a corrupted mask latent, and the semantic embedding $c^{0}$, derived from the image. Their direct fusion is impaired by the domain gap between them. SDC instead bridges them in the frequency domain across $N = 4$ blocks, each comprising two symmetric cross-attention modules. Within a module, both streams are transformed by a three-dimensional rFFT, and an affinity is formed.

\begin{equation}
\mathcal{M} = \mathrm{Softmax}\!\Big( \tfrac{1}{\tau} \big(\mathcal{F}(c^{0}) W_{q}\big)^{\!\top} \big(\mathcal{F}(e) W_{k}\big) \Big),
\label{eq:affinity}
\end{equation}

where $\mathcal{F}(\cdot)$ denotes the rFFT, $W_{q}$ and $W_{k}$ the learnable query and key projections applied in Fourier space, $\tau$ a learnable temperature, and $\mathcal{M}$ the resulting affinity map. The affinity is formed over feature \emph{channels} rather than spatial tokens and is realized as a linear-complexity cross-covariance attention over feature channels~\cite{ali2021xcit}. It renders the operation tractable in the $16^{3}$ latent resolution, whereas the channel formulation reduces this to $\mathcal{O}(NC^{2})$ for $N$ tokens and $C$ channels while preserving global mixing.

\paragraph{Timestep-adaptive band-pass filter}
The informative frequency band varies with the noise level: at high $t$ only coarse structure survives corruption, whereas at low $t$ fine boundary detail becomes recoverable. SB therefore gates the attended spectrum with a timestep-conditioned filter. To keep the gate smooth over a continuous frequency range, it is produced by a coordinate network rather than free per-bin weights: each rFFT bin is described by its normalized radial frequency $r \in [0,1]$, lifted with sinusoidal Fourier features~\cite{tancik2020fourier} and mapped through $R = 6$ FiLM-modulated blocks~\cite{perez2018film} and a sigmoid to
\begin{equation}
\mathcal{M}' = g(r,t) \odot \big( \mathcal{M} \, (\mathcal{F}(c^{0}) W_{v}) \big),
\label{eq:nbp}
\end{equation}
where $g(r,t) \in (0,1)$ is the learned gate, $W_{v}$ the value projection, $\odot$ element-wise multiplication, and $\mathcal{M}'$ the gated spectrum. Shared across channels, the gate acts as a spectral modulation whose passband tracks the noise level of step $t$. The gated spectrum is returned to Euclidean space by the inverse rFFT, added residually to the query stream, and refined by a channel MLP; the symmetric module reverses the two streams to form the next block's input, and after $N_{b}$ blocks the refined noise embedding is read by the $\epsilon$ prediction head.

\paragraph{Conditioning injection}
The clean latent $\hat{z}_{0}$ recovered from the denoiser is injected into the bottleneck through a $1^{3}$ convolution $\mathcal{P}$,
\begin{equation}
\tilde{c}^{\,0} = c^{0} + \mathcal{P}(\hat{z}_{0}),
\label{eq:inject}
\end{equation}
so that the diffusion output acts as a correction upon the semantic condition rather than as a replacement for it. The refined bottleneck $\tilde{c}^{\,0}$ is passed to ME$^2$FB and thereafter decoded to the segmentation.

\subsubsection{Multiscale Expert Enhanced Feature Block}
\label{subsec:m2}

Encoder stages at different depths encode qualitatively different information: shallow, high-resolution stages preserve local anatomical detail, whereas deeper stages capture global context. The standard SegResNet-VAE, however, constructs its bottleneck from the deepest stage alone and admits the shallower stages only as passive skip connections, so that the evidence available at the point of fusion is restricted to a single scale. ME$^2$FB removes this restriction by combining features across stages through a mixture of experts with expert-choice routing. Each of the $n$ intermediate stages is projected to a common width $d$ by a $1^{3}$ convolution and pooled to the bottleneck resolution, yielding at every bottleneck voxel a group of stage descriptors $X \in \mathbb{R}^{n \times d}$. Expert preferences are obtained from a gating function perturbed by learned noise,

\begin{equation}
\begin{aligned}
H&=g(X,\theta)^{\!\top}+R_{\text{noise}},\\
S_I&=\operatorname{Softmax}\!\left(H[\operatorname{TopK}(H,k)]\right).
\end{aligned}
\label{eq:gating}
\end{equation}

where $g(X,\theta)$ produces the expert-preference logits and $R_{\text{noise}} = \varepsilon \odot \mathrm{softplus}(X _{\text{noise}})$ promotes exploration and stabilizes optimization, and is disabled at inference so that routing is deterministic. Routing follows the expert-choice convention rather than each descriptor selecting its experts; each expert selects its own $k$ descriptors, with $k = \lfloor B \cdot V \cdot n / E \rfloor$ for $V = 16^{3}$ bottleneck voxels and $E$ experts. Every expert therefore processes an identical share of the token set, which guarantees balanced utilization by construction and precludes expert collapse without recourse to an auxiliary balancing loss. Each expert is realized as a two-layer perceptron with expansion factor four,

\begin{equation}
\hat{X}_{I} = \mathrm{ReLU}(X_{I} W_{1}) W_{2},
\qquad
W_{1} \in \mathbb{R}^{d \times 4d}, \;\; W_{2} \in \mathbb{R}^{4d \times d}
\label{eq:expert}
\end{equation}

and a descriptor selected by several experts is recombined as $X_{\text{final}} = \sum_{I} S_{I} \hat{X}_{I}$, while a residual connection carries forward unselected descriptors. The $n$ enhanced stage descriptors at each voxel are collapsed by a learnable softmax-weighted sum, refined by self-attention over the $16^{3}$ spatial tokens to inject global context, and mapped back to the bottleneck width by a $1^{3}$ convolutional neck. The result is added residually to the bottleneck, $z = \mathrm{ReLU}(z_{\text{btln}} + z_{\text{enh}})$.

\subsection{Evaluation Metrics}
\label{subsec:metrics}

Following the BraTS convention, performance is reported over four regions derived from the predicted label map: enhancing tumor (ET), tumor core (TC $=$ ET $\cup$ NCR), whole tumor (WT $=$ ET $\cup$ NCR $\cup$ ED), and resection cavity (RC), the last specific to the post-treatment setting and evaluated separately. Each region is assessed with the Dice Similarity Coefficient (DSC) and Intersection over Union (IoU), which measure volumetric overlap; sensitivity (Sn) and specificity (Sp), the true-positive and true-negative rates; and the 95th-percentile Hausdorff Distance (HD95), which measures boundary agreement and is robust to isolated surface outliers~\cite{hatamizadeh2022swinunetr}. Higher DSC, IoU, Sn, and Sp indicate better performance, whereas HD95 is reported in millimeters, with lower values preferred. All metrics are computed per case and averaged over the held-out test set. 

\setlength{\tabcolsep}{9.3pt}
\begin{table*}[t]
\centering
\renewcommand{\arraystretch}{1.4}
\caption{Performance Comparison of Single Segmentation Models on the \textbf{Test} Set for Four Regions (ET, TC, WT, RC).}
\resizebox{\textwidth}{!}{
\begin{tabular}{lccccccccccccccc}
\hline
\textbf{Models} 
& \multicolumn{5}{c}{\textbf{Dice (\%)  $\uparrow$ }} 
& \multicolumn{5}{c}{\textbf{IoU (\%)  $\uparrow$}} 
& \multicolumn{5}{c}{\textbf{HD95 (mm)  $\downarrow$}} \\
\cline{2-16}
 & \textbf{ET} & \textbf{TC} & \textbf{WT} & \textbf{NETC} & \textbf{RC} & \textbf{ET} & \textbf{TC} & \textbf{WT} & \textbf{NETC} & \textbf{RC} & \textbf{ET} & \textbf{TC} & \textbf{WT} & \textbf{NETC} & \textbf{RC} \\
\hline
\textbf{SAM} & 55.12 & 65.46 & 74.10 & 60.57&51.31 & 
46.54 & 52.50 & 60.40 & 58.10&42.26 &
9.47 & 16.06 & 15.55 & 5.78 & 18.61 \\

\textbf{UNETR} & 58.69 & 78.39 & 87.68 &51.64& 60.85
& 50.77 & 68.98 & 79.41 &47.56& 52.42 
& 6.93 & 5.96 & 4.46 & 5.49&8.99 \\

\textbf{AHNet} & 70.16 & 79.58 & 87.07 &64.75 & 63.93 
& 62.20 & 70.21 & 78.61 & 60.44&55.44 
& 5.04 & 4.82 & 3.68 &4.36& 7.42 \\

\textbf{UNet} & 63.69 & 81.84 & 89.65 &65.21 & 67.10 
& 55.81 & 73.31 & 82.24 &60.83& 58.81 
& 4.78 & 4.28 & 3.33 &3.99& 7.71 \\

\textbf{MedNeXt} & 75.11 & 84.33 & 89.19 &65.73 & 67.23 
& 67.58 & 76.48 & 81.88 & 61.36&59.02 
& 3.58 & 3.55 & 2.86 & 3.85& 6.44 \\

\textbf{SwinUNETR} & 76.72 & 85.83 & 90.77 &69.78 & 73.70 
& 69.62 & 78.22 & 83.91 &65.76& 65.68 
& 3.62 & 3.25 & 2.51 &\textbf{3.62}& 5.49 \\

\textbf{SegResNetVAE} & 77.92 & 86.70 & 90.65&69.12 & 73.10 
& 70.86 & 79.18 & 83.86 &65.03& 64.97
& 3.59 & 3.33 & 2.45 &3.98& 5.91 \\

\textbf{ME$^2$FDiff} & \textbf{78.75} & \textbf{87.14} & \textbf{91.23} & \textbf{70.08} &\textbf{74.20} & \textbf{71.60} & \textbf{79.71} & \textbf{84.52} &\textbf{65.92}& \textbf{66.62} & \textbf{3.56} & \textbf{3.24} & \textbf{2.13}& 3.87 & \textbf{5.47} \\
\hline
\end{tabular}
}

\label{table:1}
\end{table*}

\begin{figure*}[h!]
    \centering
    \includegraphics[width=1\linewidth]{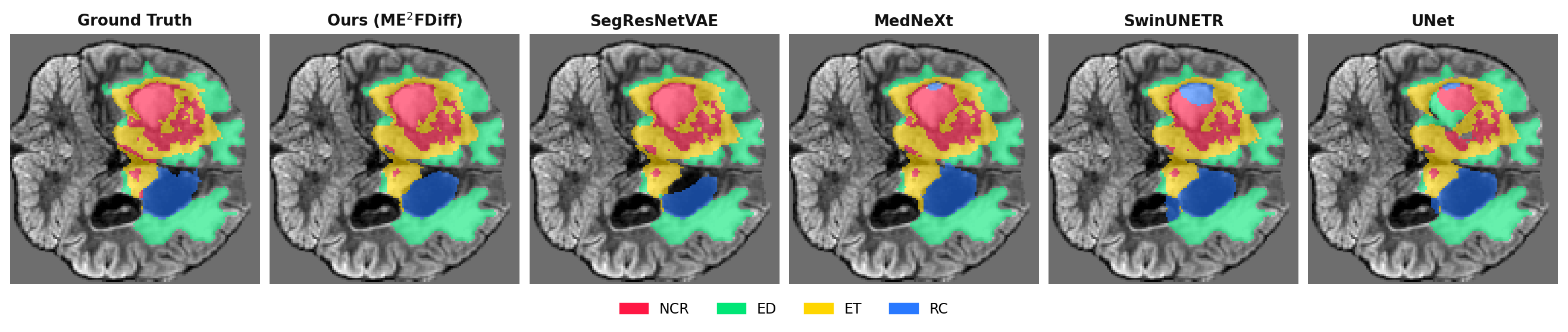}
    \caption{Qualitative comparison among the predicted masks for baseline models with ground truth.}
    \label{figure:3}
\end{figure*}


\section{Experimental Setup} \label{sec:experimental_setup}
\subsection{Training configuration}
\label{subsec:adamw}
All models are optimized using AdamW with a peak learning rate of $1 \times 10^{-4}$ \cite{loshchilov2019decoupled} and a minimum learning rate of $1 \times 10^{-6}$, following a cosine annealing schedule. Models are trained for a maximum of $150$ epochs with early stopping (patience of $40$ epochs, monitored on validation Dice). Training uses a batch size of $1$ per GPU under automatic mixed-precision (AMP) with FP16, distributed across $4$ GPUs (NVIDIA A6000/A100 of Rivanna/Afton high-performance computing system) using PyTorch Distributed Data Parallel (DDP). A global random seed of $42$ is used, with each distributed rank offset by its rank index ($\text{seed} + \text{rank}$) to ensure reproducible yet decorrelated data sampling across processes.

\setlength{\tabcolsep}{7.2pt}
\begin{table*}[]
\centering
\renewcommand{\arraystretch}{1.4}
\caption{Performance Comparison of ensemble fusion strategies with three models (SegResNetVAE + SwinUNETR + MedNeXt) on the \textbf{Test} set for four regions (ET, TC, WT, RC) with corresponding training time (TT) per epoch.}
\resizebox{\textwidth}{!}{
\begin{tabular}{lcccccccccccccccc}
\hline
\textbf{Models} &
\textbf{TT (mm:ss) $\downarrow$} &
\multicolumn{5}{c}{\textbf{Dice (\%) $\uparrow$}} &
\multicolumn{5}{c}{\textbf{IoU (\%) $\uparrow$}} &
\multicolumn{5}{c}{\textbf{HD95 (mm) $\downarrow$}} \\
\cline{3-17}
 && \textbf{ET} & \textbf{TC} & \textbf{WT} & \textbf{NETC} & \textbf{RC} & \textbf{ET} & \textbf{TC} & \textbf{WT} & \textbf{NETC} & \textbf{RC} & \textbf{ET} & \textbf{TC} & \textbf{WT} & \textbf{NETC} & \textbf{RC} \\
\hline
\textbf{Dynamic Weighting} & 26:89 & 67.76 & 84.34 & 90.66 &70.01 & 68.29 & 60.41 & 76.65 & 83.76 &63.48& 60.09 & 4.18 & 3.01 & 3.01&3.53 & 7.06 \\

\textbf{Cascade Fusion} & 30:56 & 78.68 & 85.11 & 90.77 &65.19& 65.54 & 71.65 & 77.53 & 83.91&60.94 & 57.48 & \textbf{3.04} & 3.15 & 2.52 &3.58& 6.88 \\

\textbf{MoE}& 39:87 & 77.59 & 86.45 & 91.18 &68.79& 73.31 & 70.63 & 79.09 & 83.59 &64.22& 66.36 & 3.20 & \textbf{2.92} & 2.17 & \textbf{3.34}&\textbf{5.21} \\

\textbf{ME$^2$FDiff}& \textbf{10:92}  & \textbf{78.75} & \textbf{87.14} & \textbf{91.23} & \textbf{70.08} &\textbf{74.20} & \textbf{71.60} & \textbf{79.71} & \textbf{84.52} &\textbf{65.92}& \textbf{66.62} & \textbf{3.56} & \textbf{3.24} & \textbf{2.13}& 3.87 & 5.47 \\
\hline
\end{tabular}
}
\label{table:2}
\end{table*}
\begin{figure*}[h!]
    \centering
    \includegraphics[width=1\linewidth]{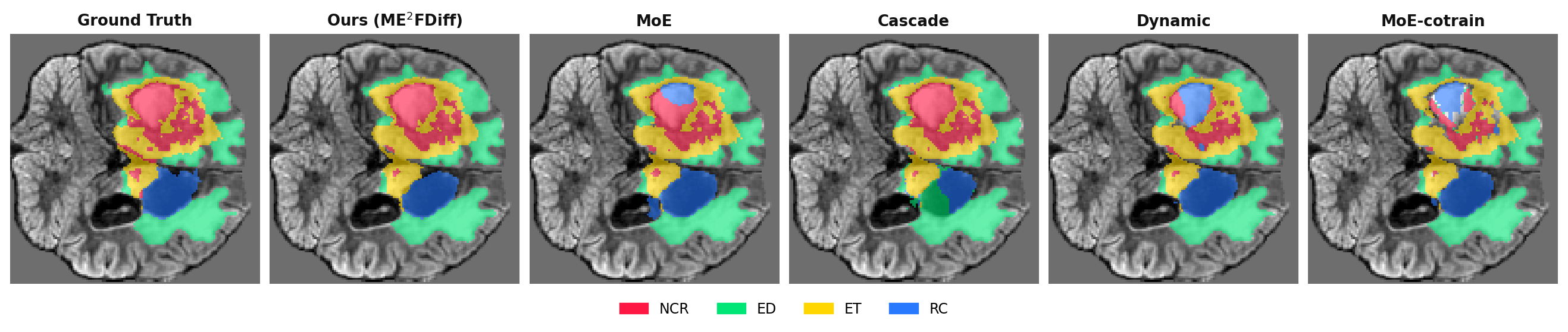}
    \caption{Qualitative comparison among the predicted mask for ensemble models with ground truth.}
    \label{figure:4}
\end{figure*}


\subsection{Loss configuration}
To handle the complexity of different regions and their imbalance ratios, all models are trained with a compound loss combining three loss functions, i.e., Tversky~\cite{salehi2017tversky},  Dice~\cite{milletari2016vnet}, and Focal terms~\cite{lin2017focal}.
\begin{equation}
\mathcal{L}_{\text{total}} = 0.4\,\mathcal{L}_{\text{Tversky}} + 0.3\,\mathcal{L}_{\text{Dice}} + 0.3\,\mathcal{L}_{\text{Focal}}.
\label{eq:compound_loss}
\end{equation} 
where Dice provides stable overlap optimization, Tversky accounts for recall bias in small critical regions, and Focal adds voxel-level attention to hard boundaries. When combined, they operate at different granularities and cover each other's gaps. The class weighting encodes the same priority. Rather than strict inverse-frequency weighting, which destabilizes training when a region is absent from a volume, weights are assigned in inverse rank of expected sub-region volume and confined to a narrow range: $0.5$ for background, $1.0$ for the largest tumor region (SNFH), $2.0$ for NETC, and $2.5$ for the small ET and RC. A false-negative penalty $\beta = 0.8$ is applied to the small ET and RC
regions, biasing the Tversky term toward recall where a missed lesion is costlier than an enlarged boundary, and the Focal term uses $\gamma = 2.0$ and $\alpha = 0.25$, concentrating the gradient on the ambiguous voxels that dominate post-treatment boundaries.

\section{Results}
\label{sec:results}

We evaluate the proposed model against progressively stronger points of comparison. Section~\ref{subsec:results_single} reports performance against individually trained segmentation architectures; Section~\ref{subsec:results_ensemble} against ensemble fusion strategies that aggregate several such architectures and therefore constitute the more demanding baseline; and Section~\ref{subsec:results_sota} against recent state-of-the-art methods, considering accuracy alongside model size. It also establishes the statistical reliability of the observed differences, and Section~\ref{subsec:ablation_components} isolates the contribution of each architectural component. All results are reported on the held-out test set, with Dice, IoU, Sensitivity, and HD95 computed independently for the four tumor sub-regions (ET, TC, WT, and RC) as defined in Section~\ref{subsec:metrics}.

\subsection{Comparison with Single Model Baselines}
\label{subsec:results_single}

Table~\ref{table:1} reports test-set performance across the four tumor sub-regions. The SAM-based foundation models perform worst (SAM: $61.50$ mean Dice), reflecting the difficulty of transferring 2D natural-image pretraining to volumetric MRI, while SwinUNETR ($81.76$) and SegResNetVAE ($82.09$) are the strongest baselines. The proposed model attains the highest Dice and IoU on every sub-region, improving over SegResNetVAE by $0.83$, $0.44$, $0.58$, and $1.10$ Dice points on ET, TC, WT, and RC, and raising mean Dice from $82.09$ to $82.83$. The largest gain is on RC, the hardest region for all baselines and the one unique to the post-treatment setting. It also achieves the best mean HD95
($3.60$~mm), improving on SegResNetVAE by $0.03$-$0.44$~mm across the four regions, with SwinUNETR the only close competitor on boundary localization (TC $3.25$ vs.\ $3.24$~mm, RC $5.49$ vs.\ $5.47$~mm). As the overlap gains are consistent across all regions and both overlap metrics whereas the HD95 differences are small and inconsistent in direction, the improvement is attributable principally to more accurate delineation of region extent rather than sharper boundary placement.  The qualitative performance comparison for single models is illustrated in Figure \ref{figure:3}.

\subsection{Comparison with Ensemble Baselines}
\label{subsec:results_ensemble}

Table~\ref{table:2} compares the proposed model against four ensemble fusion strategies, each aggregating the same three best-performing architectures (SegResNetVAE, SwinUNETR, MedNeXt) shown in Table~\ref{table:1}. Among the ensembles, learned gating is clearly superior to the alternatives: MoE attains the best mean Dice ($82.13$), ahead of Cascade Fusion ($80.03$) and Dynamic Weighting ($77.76$). The margin is most pronounced on RC, where MoE reaches $73.31$ Dice against $65.54$ for Cascade, evidence that confidence-derived and sequentially stitched fusion rules both degrade sharply on the hardest sub-region, whereas a learned gate does not. Crucially, the proposed model surpasses the strongest ensemble on both overlap metrics while requiring only a single forward pass. It exceeds MoE on mean Dice ($82.83$ vs.\ $82.13$) and mean IoU, winning three of four sub-regions on each; the remaining case (WT for IoU) differs by only $0.07$ points and is effectively tied. That a single network matches and exceeds a three-expert ensemble on volumetric overlap at approximately one-third of its inference cost. It indicates that the expert specialization such ensembles achieve through external fusion can instead be internalized within a single architecture. Each ensemble must train three separate expert networks together with its fusion stage, taking between $26.89$ and $39.87$~minutes (dynamic $26.89$, cascade $30.56$, MoE $39.87$), whereas the proposed model is a single network and trains in $10.92$~minutes-a $2.5$-$3.7\times$ reduction. The saving carries over to deployment: inference requires a single forward pass rather than the three needed to evaluate every expert. The proposed model thus trades a small amount of boundary regularity for stronger region delineation at a fraction of the training and inference cost, a favorable trade-off where accurate estimation of tumor extent, rather than sub-millimeter boundary placement, drives clinical utility. The qualitative performance comparison for ensemble models is illustrated in Figure \ref{figure:4}.

\begin{table}[ht]
\centering
\renewcommand{\arraystretch}{1.3}
\caption{Performance comparison of existing SOTA segmentation models for the whole tumor on Test test. Parameters are reported as Millions (M) and Dice, IOU, Acc, Sn, Sp are reported as percentages (\%).}
\resizebox{\columnwidth}{!}{
\label{tab:model_comparison}
\begin{tabular}{lcccccc}
\hline
\textbf{Model} &\textbf{params} & \textbf{Dice} & \textbf{IOU} & \textbf{Acc}&\textbf{Sn}& \textbf{Sp}\\
\hline
MoE-SAM& 134.29&79.74 & 67.60 & 99.20 & 80.47 & 99.58 \\

SegFormer& 42.67&85.14 & 76.11 & 99.48 & 85.17 & 99.75 \\

TransUNet&132.39& 90.08 & 82.87 & 99.63 & 90.64 & 99.77 \\

MedFormer&39.59& 90.57 & 83.76 & 99.67 & 89.95 & \textbf{99.85}\\

UMamba &31.84& 90.92 & 84.14 & 99.66 &91.03 &99.83\\

\textbf{ME$^2$FDiff } & \textbf{28.82} &\textbf{91.23} & \textbf{84.65} & \textbf{99.67} & \textbf{91.24} & 99.83 \\
\hline
\end{tabular}}\label{table:3}
\end{table}

\begin{figure}[h!]
    \centering
    \includegraphics[width=1\linewidth]{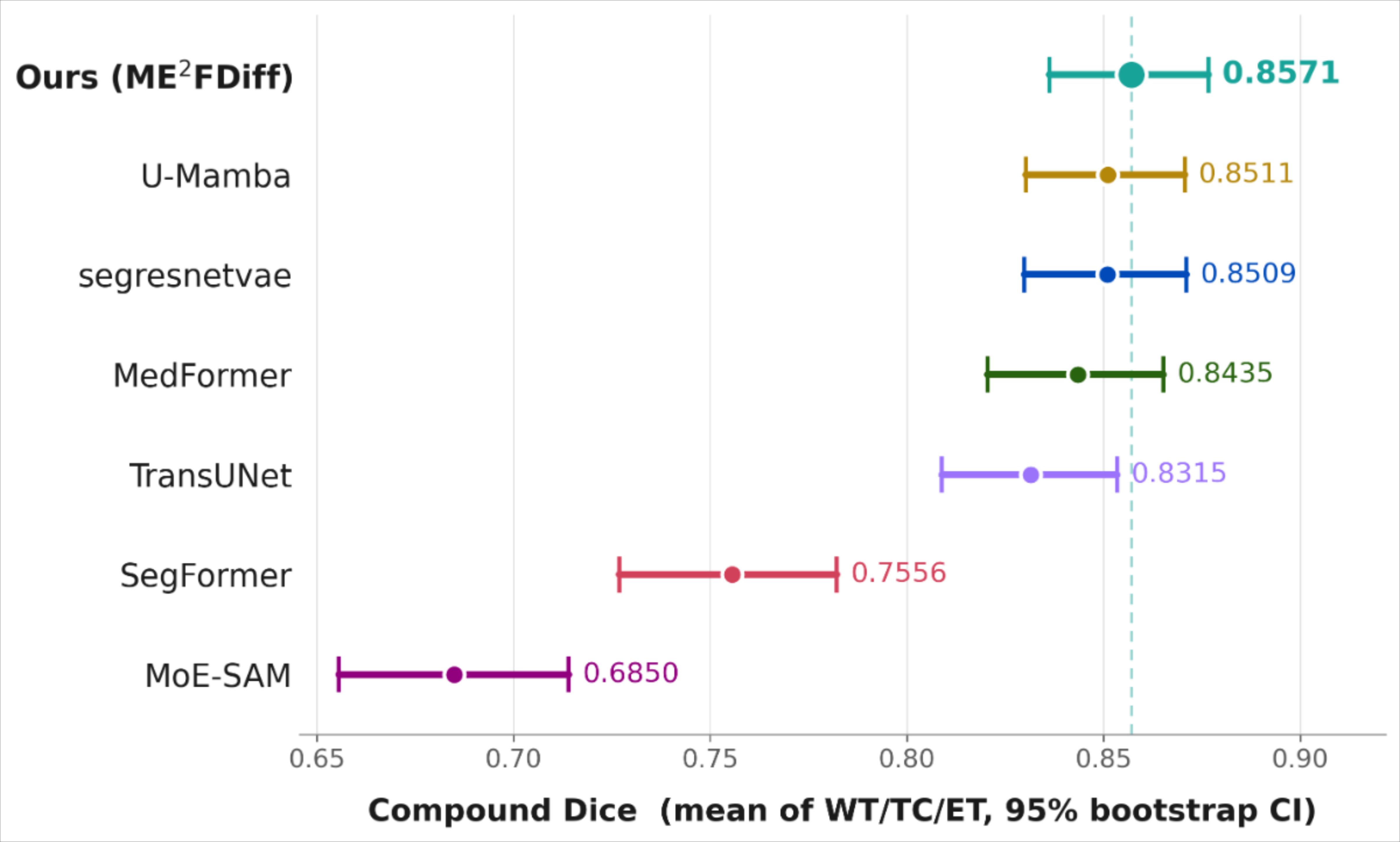}
    \caption{Compound Dice of the mean of the three BraTS tumor sub-regions (whole tumor WT, tumor core TC, enhancing tumor ET) on the held-out test set (163 cases). Each marker is the per-method mean and the horizontal bar its 95\% bootstrap confidence interval (5,000 resamples); the dashed vertical line marks our mean for reference. All competing methods were retrained on the identical data split and training criteria, so differences reflect the model, not the pipeline.}
    \label{figure:5}
\end{figure}
\subsection{Comparison with State-of-the-Art Methods}
\label{subsec:results_sota}

We compare the proposed model against recent state-of-the-art segmentation architectures. To ensure that observed differences reflect the model rather than the training pipeline, all competing methods were retrained under identical data splits and training criteria. Table~\ref{table:3} reports whole-tumor performance alongside parameter count, and Figure.~\ref{figure:5} shows the compound Dice over the three BraTS sub-regions (WT, TC, ET) on the held-out test set (163 cases), with 95\% bootstrap confidence intervals. The proposed model attains the best Dice (91.23), IoU (84.65), and Sensitivity (91.24) of all methods evaluated, while using the \emph{fewest} parameters (28.82M). It exceeds the strongest competing method, UMamba, by 0.31 Dice and 0.51 IoU with 9.5\% fewer parameters and surpasses the transformer-based TransUNet by 1.15 Dice using 4.6$\times$ fewer parameters.

The margin over MoE-SAM is largest (+11.49 Dice at 4.7$\times$ fewer parameters), consistent with the limited transferability of natural-image foundation models to volumetric MRI. Accuracy and specificity are effectively saturated across all competitive methods and therefore carry little discriminative signal. Figure.~\ref{figure:5} confirms that this ordering is preserved when performance is aggregated across all three tumor sub-regions rather than assessed on the whole tumor alone. Across the comparison, accuracy and model size move together in the proposed model's favor: it records the highest overlap accuracy while carrying the fewest parameters of any method evaluated. This is notable given the prevailing trend toward increasingly large architectures: the two largest models in the comparison, MoE-SAM (134.29M) and TransUNet (132.39M), are both outperformed by a model roughly one-fifth their size, supporting the view that architectural specialization, rather than raw capacity, is the principal determinant of performance on this task.

\begin{table}[ht]
\centering
\renewcommand{\arraystretch}{1.3}
\caption{Component-wise Overall Contribution to the whole tumor region for our proposed method. Dice, IOU, and Sn are reported as percentages (\%), and HD95 is reported as Milimeters(mm)}
\resizebox{\columnwidth}{!}{
\begin{tabular}{p{4cm}cccc}
\hline
\textbf{Configuration} & \textbf{Dice} & \textbf{IOU} & \textbf{Sn} & \textbf{HD95}\\
\hline
SegResNetVAE (Baseline)& 90.66 & 83.86 &89.74 &2.45 \\

+ Multiscale Expert Enhanced Feature Block (ME$^2$FB) & 90.81 &84.07 &89.91 &2.42\\

+ Attention-Conditioned Spectral Diffusion (ACSDiff) & 91.23 &84.52 &91.80&2.13\\
\hline
\end{tabular}}
\label{table:4}
\end{table}
\subsection{Component Contribution Analysis}
\label{subsec:ablation_components}

\begin{figure}[h!]
    \centering
    \includegraphics[width=1\linewidth]{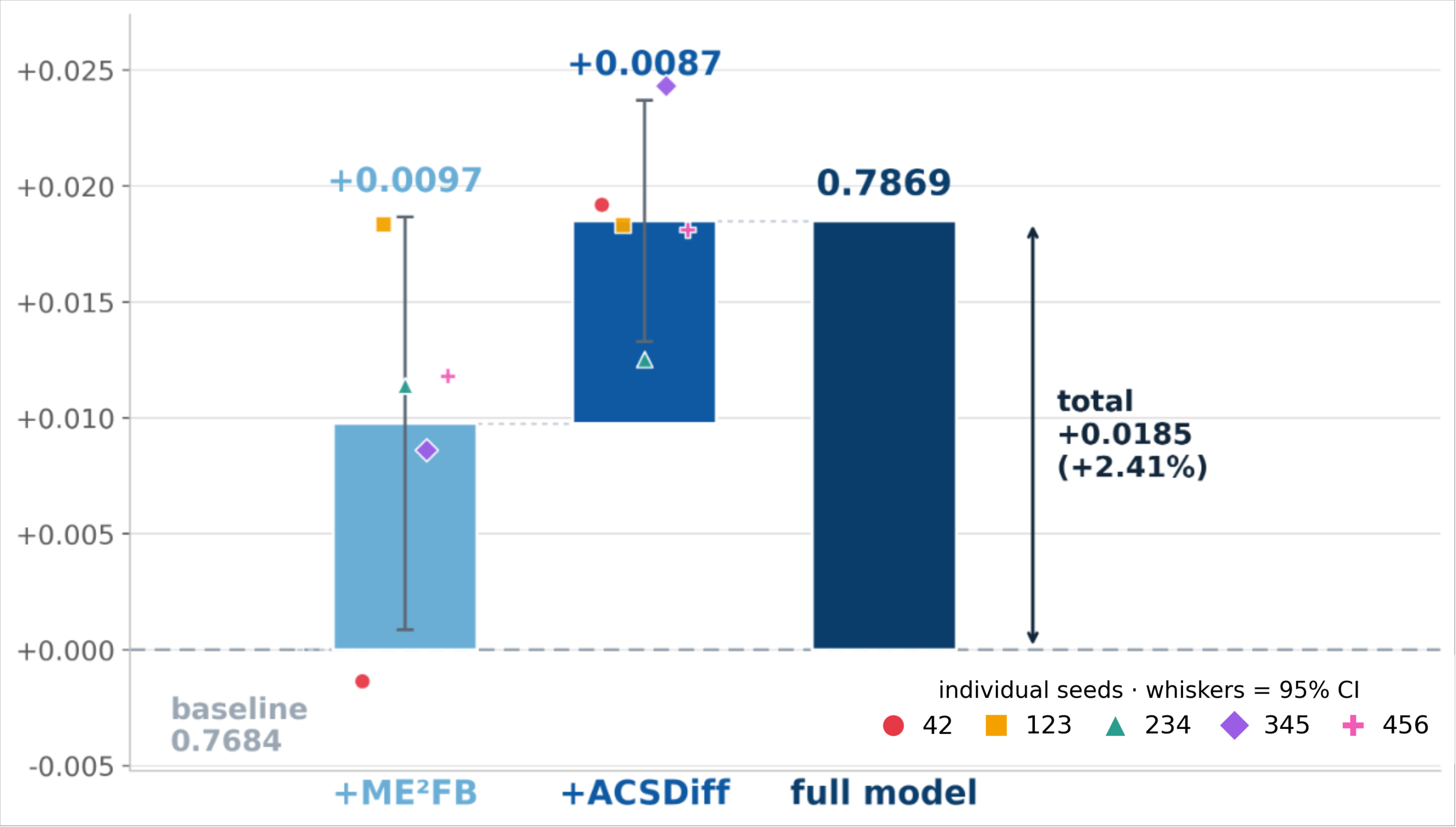}
    \caption{Component-wise contribution to Enhancing-Tumor (ET) Dice. Waterfall decomposition over the SegResNetVAE baseline. Bar heights are the cumulative gain over baseline; the bold value above each bar is the incremental improvement of that step, averaged over the training seeds. Whiskers are 95\% confidence intervals of the mean and colored markers show individual seeds.}
    \label{figure:6}
\end{figure}

To isolate the contribution of each architectural component, we add modules successively to the SegResNetVAE baseline. Table~\ref{table:4} reports whole-tumor (WT) performance, and Figure.~\ref{figure:6} decomposes the corresponding gain on enhancing tumor (ET) over five different seeds. We report ET separately as it delineates the actively proliferating tumor that defines the treatment target volume and the primary criterion for response assessment, and is the hardest region to segment in the post-treatment setting, where genuine tumor enhancement must be distinguished from treatment-related enhancement. On WT, each component improves every metric monotonically. The ME$^2$FB raises Dice from $90.66$ to $90.81$ and the ACSDiff by a further $0.42$ point, giving $91.23$ Dice and $84.52$ IoU. The gain is most pronounced in Sensitivity, which rises by $2.06$ points overall ($89.74 \rightarrow 91.80$), indicating that the added components principally recover tumor tissue missed by the baseline, while HD95 falls from $2.45$ to $2.13$~mm.

The same ordering holds on ET (Figure.~\ref{figure:6}). From an ET Dice of $0.7684$, ME$^2$FB contributes $+0.0087$, and the ACSDiff reported, inclusive of SPG, is consistent with Table~\ref{table:4}, a further $+0.0097$, reaching $0.7869$: a cumulative gain of $+0.0185$, or $+2.41\%$ relative to the baseline. The per-seed values indicate that both components improve ET Dice on average rather than uniformly across seeds.

\section{Conclusion}
Ensembling remains the default answer to architecture-dependent variability in tumor segmentation, at the price of several networks and a combination rule fixed in advance. ME$^2$FDiff shows the price is avoidable: routing applied across encoder scales supplies the complementarity an ensemble buys with extra models, and refinement applied to a mask latent supplies the correction step a single forward pass otherwise lacks. On BraTS 2024 post-treatment glioma, the result is a model that is simultaneously the most accurate and the smallest we evaluated. Where it still falls short is instructive. The resection cavity, the new sub-region introduced by the post-treatment setting, resists every method tested, ours included, suggesting the remaining difficulty is one of training signals rather than architecture. Because ACSDiff's reverse process is stochastic, sampling it repeatedly yields multiple hypotheses at no architectural cost, an avenue we leave open alongside extension to the wider BraTS 2024 cluster.

\bibliographystyle{IEEEtran}
\bibliography{ref}

\end{document}